\documentclass[10pt,twocolumn,letterpaper]{article}

\usepackage[pagenumbers]{cvpr} % To force page numbers, e.g. for an arXiv version

\usepackage{graphicx}
\usepackage{amsmath}
\usepackage{amssymb}
\usepackage{booktabs}
\usepackage{algorithm}
\usepackage{algorithmicx}
\usepackage{multirow}
\usepackage{algpseudocode}
\usepackage{indentfirst}
\usepackage{array}
\usepackage{color}

\DeclareMathOperator*{\argmin}{arg\,min}
\usepackage{caption}
\usepackage[pagebackref,breaklinks,colorlinks]{hyperref}

\definecolor{mypurple}{RGB}{143,78,151}

\usepackage[capitalize]{cleveref}
\Crefname{section}{Sec.}{Secs.}
\Crefname{section}{Section}{Sections}
\Crefname{table}{Table}{Tables}
\Crefname{table}{Tab.}{Tabs.}

\def\cvprPaperID{317} % *** Enter the CVPR Paper ID here
\def\confName{CVPR}
\def\confYear{2023}

\begin{document}

\title{Downscaled Representation Matters: Improving Image Rescaling with Collaborative Downscaled Images}

\author{Bingna Xu\footnotemark[1], Yong Guo\footnotemark[1], Luoqian Jiang, Mianjie Yu, Jian Chen\footnotemark[2]\\
South China University of Technology\\
{\tt\small sexbn@mail.scut.edu.cn, guoyongcs@gmail.com,}\\
{\tt\small \{seluoqianjiang,202030482362\}@mail.scut.edu.cn, ellachen@scut.edu.cn}
% \and
}

\maketitle

\renewcommand{\thefootnote}{\fnsymbol{footnote}}
\footnotetext[1]{Authors contributed equally.}
\footnotetext[2]{Corresponding author.}
%%%%%%%%% ABSTRACT
\begin{abstract}

Deep networks have achieved great success in image rescaling (IR) task that seeks to learn the optimal downscaled representations, i.e., low-resolution (LR) images, to reconstruct the original high-resolution (HR) images. Compared with super-resolution methods that consider a fixed downscaling scheme, e.g., bicubic, IR often achieves significantly better reconstruction performance thanks to the learned downscaled representations. This highlights the importance of a good downscaled representation in image reconstruction tasks. Existing IR methods mainly learn the downscaled representation by jointly optimizing the downscaling and upscaling models. Unlike them, we seek to improve the downscaled representation through a different and more direct way -- optimizing the downscaled image itself instead of the down-/upscaling models. Specifically, we propose a collaborative downscaling scheme that directly generates the collaborative LR examples by descending the gradient w.r.t. the reconstruction loss on them to benefit the IR process. Furthermore, since LR images are downscaled from the corresponding HR images, one can also improve the downscaled representation if we have a better representation in the HR domain. Inspired by this, we propose a \textbf{Hierarchical Collaborative Downscaling (HCD)} method that performs gradient descent in both HR and LR domains to improve the downscaled representations. Extensive experiments show that our HCD significantly improves the reconstruction performance both quantitatively and qualitatively. Moreover, we also highlight the flexibility of our HCD since it can generalize well across diverse IR models.
% Image downscaling and image upscaling are often viewed as closely related tasks. In order to save data transmission costs, the original high-resolution image is usually downscaled, and in some specific scenarios, it has to be restored to the original resolution. To maximize the restoration performance, some advanced work jointly trains downscaling network and the upscaling network. Therefore, we observe that the manifestations of the LR image have a greater contribution to the performance of the upscaling model. To this end, We propose a collaborative example generation method based on image reconstruction with no training. More specifically, we directly optimize LR images under the supervision of real HR to make them more suitable for upscaling models. To further improve the performance, we also propose a Hierarchical Collaborative Downscaling (HCD)  scheme, which first optimizes the HR image, and then optimizes the LR image obtained by HR downscaling. Experimental results show that our model significantly outperforms existing methods in both quantitative and qualitative evaluations of image upscale reconstruction from downscaled images.
\end{abstract}

%%%%%%%%% BODY TEXT
\section{Introduction}
\label{sec:intro}

% $*$\footnote{Authors contributed equally.}
% $\dag$\footnote{Corresponding author.}
% Par1-image rescaling task definition and application
% \yong{Image Rescaling (IR) seeks to xxx}
% Image rescaling seeks to downscale the original high resolution (HR) images to visually-pleasing low resolution (LR) images for saving data transmission costs, and then upscale these LR images to recover the realistic HR images. It has been widely used in real-world scenarios to fit display screens with different resolutions for image preview. When a user zooms in to see more details, these images are recovered to their original high resolution on mobile devices, which has a high research value and application potential~\cite{xiao2020invertible,pan2022towards,Liang2021HierarchicalCF}.
\begin{figure}[t]
    \centering
    % \hspace{9mm}
    \centerline{\includegraphics[width=1\columnwidth]{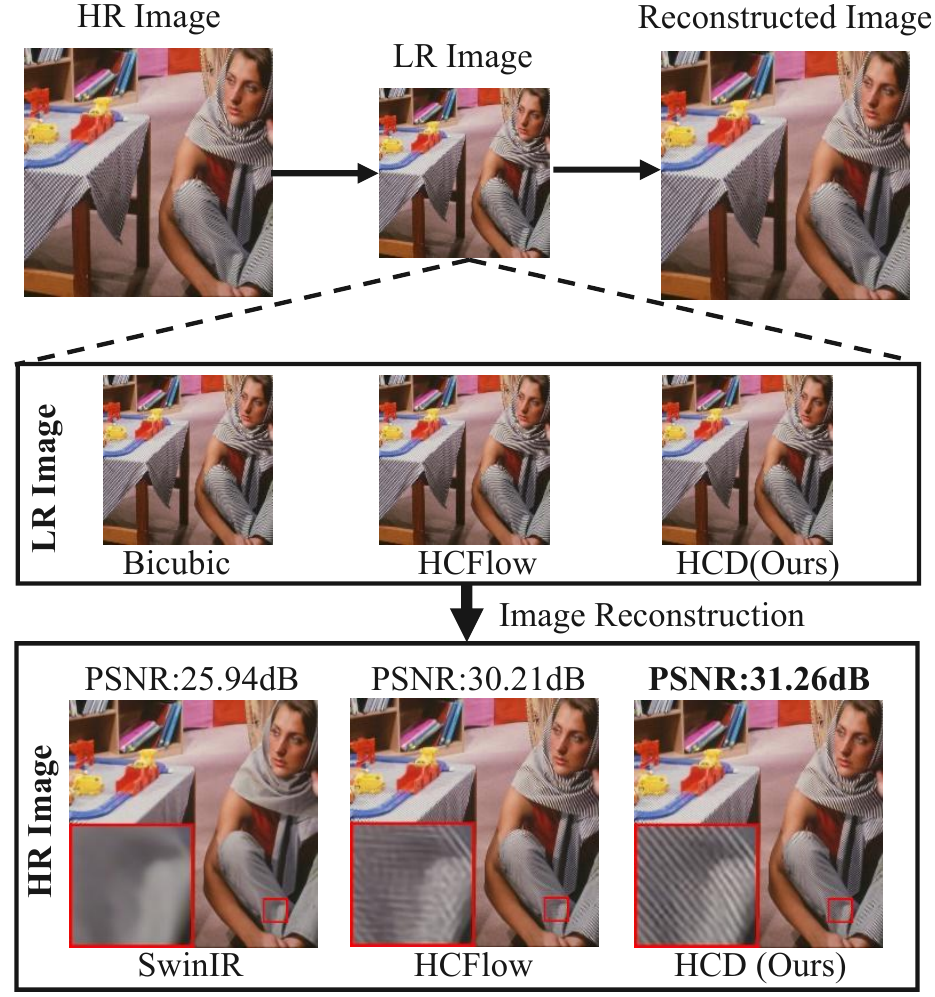}}
    % \vspace{-10 pt}
    \caption{Image rescaling pipeline and the comparisons of the downscaled images along with the corresponding reconstructed HR images (4$\times$). Top: we show the entire process of image rescaling. Middle: we visualize the downscaled representations used in different methods. Bottom: we compare the reconstructed HR images. With the improved downscaled representation, our method yields the best result both quantitatively and qualitatively.}
    \label{constrast}
%    \vspace{-10 pt}
\end{figure} 

Deep neural networks have achieved great success in many computer vision tasks, including image classification~\cite{simonyan2014very,guo2018double,guo2020multi,he2016deep,guo2021towards,guo2021content}, semantic segmentation~\cite{long2015fully,liu2020dynamic,wang2018understanding}, and many other areas~\cite{deng2022boosting,guo2019nat,niu2021adaxpert,guo2020breaking,pan2022improving}.
Recently, image rescaling has become an important task that seeks to downscale the high-resolution (HR) images to visually valid low-resolution (LR) images and then upscale them to recover the original HR images. 
In practice, the downscaled images play an important role in saving storage or bandwidth and fitting the screens with different resolutions~\cite{xiao2020invertible}, such as image/video compression and video communication~\cite{zhang2018image,sun2020learned,pan2022towards}.
% coming with a wide range of real-world applications
Interestingly, unlike super-resolution (SR)~\cite{saharia2022image,lu2022transformer,chen2022real} methods that consider a fixed downscaling kernel, e.g., bicubic, IR often yields significantly better reconstruction performance~\cite{zhu2022high,chen2021direct,gupta2021rice} since IR essentially learns a better downscaled representation method. As shown in ~\Cref{constrast}, compared with a popular SR method SwinIR~\cite{liang2021swinir}, the rescaling method HCFlow~\cite{Liang2021HierarchicalCF} greatly improves the PSNR score from 25.94 dB to 30.21 dB, which highlights the importance of a \emph{good downscaled representation} in image reconstruction tasks.

\begin{figure}[t]
    \centering
    \hspace{-3mm}
    \includegraphics[width=\columnwidth]{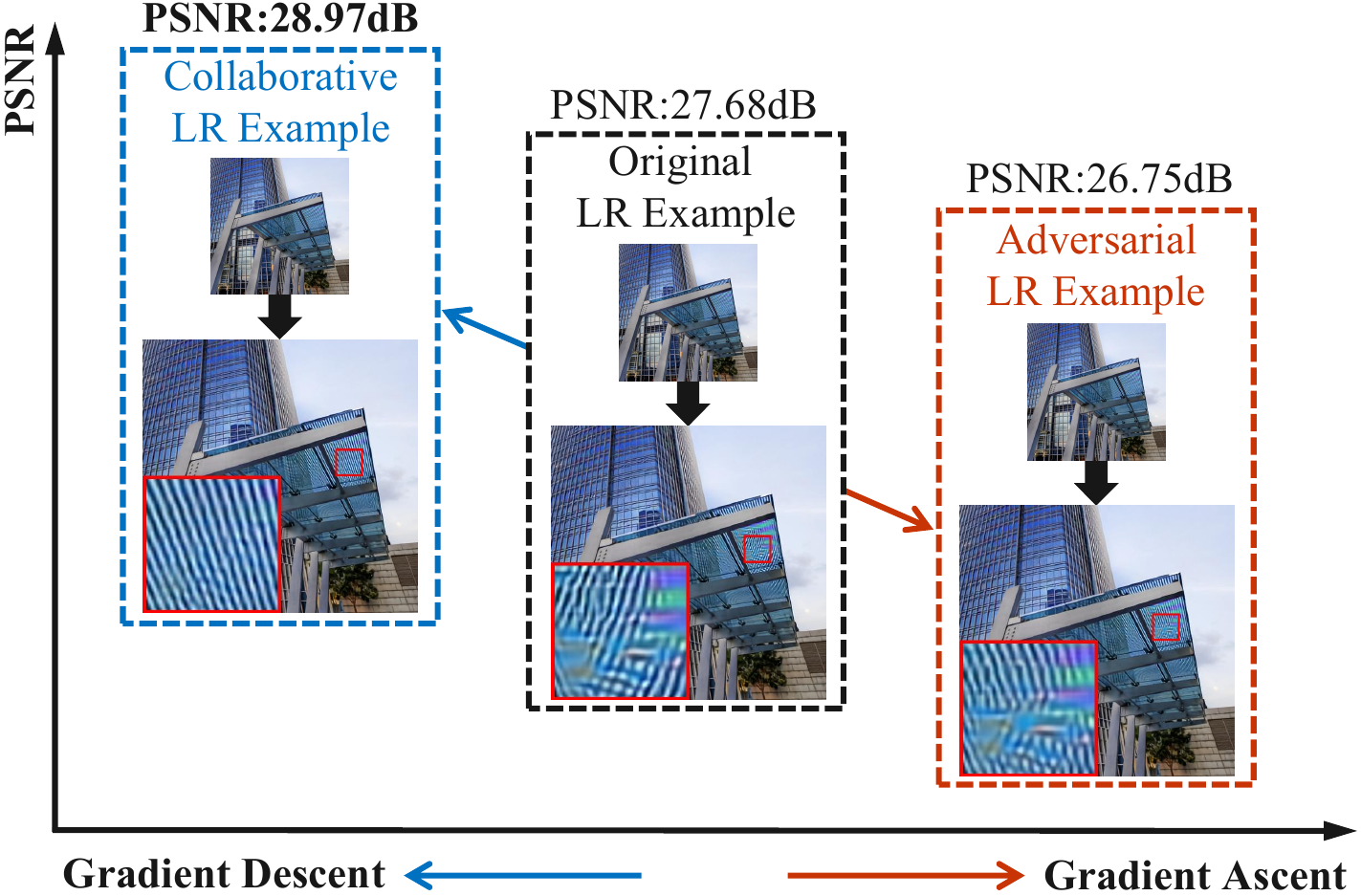}
    \caption{Comparison of adversarial examples and collaborative examples when producing downscaled representations. We ascend the gradient w.r.t the reconstruction loss to generate the adversarial examples. Based on the pretrained HCFlow model~\cite{Liang2021HierarchicalCF}, the adversarial LR example yields lower PSNR along with distorted visual content in the reconstructed HR image. By contrast, we generate the collaborative LR example by descending the gradient and obtain significantly higher PSNR as well as better visual quality.}
    \label{col vs adv}
%    \vspace{-10 pt}
\end{figure}

To learn a good downscaled representation, existing IR methods jointly learn the downscaling and upscaling models by minimizing the reconstruction loss~\cite{xiao2020invertible,Liang2021HierarchicalCF}. Besides this line of research, one can also directly optimize the downscaled representation instead of updating the model parameters. From this point of view, a typical approach is adversarial attack that iteratively optimizes the input image itself to mislead the model~\cite{madry2017towards}. Interestingly, adversarial attack is also very effective in image rescaling tasks. As shown in ~\Cref{col vs adv}, based on a popular IR model HCFlow~\cite{Liang2021HierarchicalCF}, we apply gradient ascent on LR images to generate the adversarial example. In practice, it greatly hampers the reconstruction performance by reducing the PSNR by 0.93 dB and also introduces severe visual artifacts in the reconstructed HR image. The degraded performance encourages us to explore the possibility of improving reconstruction performance -- optimizing the downscaled representation in an opposite way to adversarial attack.

Inspired by this, we propose a collaborative downscaling scheme that descends the gradient, i.e., reducing the reconstruction loss, to construct the collaborative LR example. As shown in ~\Cref{col vs adv}, the collaborative example yields a significant performance improvement of 1.29 dB along with better visual details in the reconstructed image. Furthermore, since the LR image is obtained from the original HR image via downscaling, we can also improve the downscaled representation if we obtain a better representation in the HR domain, i.e., generating a collaborative HR example. Motivated by this, we propose a \textbf{Hierarchical Collaborative Downscaling (HCD)} scheme that optimizes the representations in both HR and LR domains to obtain a better downscaled example. Specifically, we first generate a collaborative example in the HR domain and use the downscaling model to obtain a downscaled image. Taking the downscaled image as an initialization point, we then generate the collaborative LR example to further improve the downscaled representation. Due to the dependence between HR and LR image (based on the downscaling process), the hierarchical collaborative learning scheme can be formulated by a bi-level optimization problem. In practice, we solve it by alternatively optimizing the representation in HR and LR domains. More critically, we highlight that the proposed HCD does not introduce extra computational overhead in the upscaling stage, making our method applicable to real-world 
scenarios. Extensive experiments show that our method greatly boosts the reconstruction performance with the help of the collaborative downscaled examples.

Our contributions are summarized as follows:
\begin{itemize}
    \item We propose a novel collaborative image downscaling method that improves the image rescaling performance from a new perspective -- learning a better downscaled representation. We highlight that, in the community of image reconstruction, it is the first attempt to directly optimize the downscaled representation instead of learning the downscaling or upscaling models to boost the performance of image rescaling. 
    % The proposed HCD can be easily integrated with existing advanced image rescaling models.
% \item We propose a hierarchical collaborative example generation algorithm to generalize the rescaling process to a bi-level optimization problem by jointly optimizing the primeval input, i.e., the ground-truth HR image as well as the corresponding downscaled image.
    \item Since the low-resolution (LR) images strongly depend on the corresponding high-resolution (HR) images, we propose a Hierarchical Collaborative Downscaling (HCD) that optimizes the representations in both HR and LR domains to learn a better downscaled representation. We formulate the learning process as a bi-level optimization problem and solve it by alternatively generating collaborative HR and LR examples.
    
    \item Experiments on multiple benchmark datasets show that, on top of state-of-the-art image rescaling models, our HCD yields significantly better results both quantitatively and qualitatively. For example, based on a strong baseline HCFlow~\cite{Liang2021HierarchicalCF}, we obtain a large PSNR improvement of 0.7 dB on Set5 for 4$\times$ rescaling.
\end{itemize}

%-------------------------------------------------------------------------
\section{Related Work}
\label{sec:relatedwork} 
\subsection{Image Rescaling}
% In the real world, image downscaling followed by image upscaling is a common strategy for reducing image storage and transmission bandwidth costs. Previous work frequently treats the two processes mentioned above as separate tasks, namely image downscaling and image upscaling.
%  is to generate the HR version of an existing LR image, which

Image rescaling (IR) and image super-resolution (SR)~\cite{tong2017image,zhang2018learning,zhang2019deep,dai2019second} are distinct tasks. IR consists of image downscaling and image upscaling. SR corresponds to the latter process, which lacks a ground-truth HR image or any other prior information, and the reconstruction process is entirely based on the LR image. In contrast, we are given a ground-truth HR image in the IR task, but we must use its downscaled version for storage and transmission. We recover the original HR image when necessary. Image downscaling is the inverse of SR which generates the LR version of an HR image.  The most common downscaling method is bicubic interpolation~\cite{mitchell1988reconstruction}. However, it will cause over-smoothed issues since the high-frequency details are suppressed. Furthermore, most image downscaling algorithms focus solely on the visual quality of the downscaled images, which may not be suitable for the upscaling tasks. 
% For upscaling, \yu{many excellent neural SR models and methods have been
% proposed\cite{kim2016deeply,ruan2022efficient,tong2017image,zhang2018learning,zhang2019deep,dai2019second,johnson2016perceptual,glasner2009super,o2022pear}. More
% reviews and discussion about single image SR using deep learning can be referred to \cite{yang2019deep}.}

Recently, an increasing amount of work has been devoted to modeling image downscaling and upscaling as a unified task~\cite{guo2022invertible,pan2022learning,li2021approaching,hou2018learning,zhong2022faithful,wang2022cnn}. Sun et al.~\cite{sun2020learned} propose a content-adaptive resampler to achieve image downscaling and improves the upscaling model. Xiao et al.~\cite{xiao2020invertible} propose a bidirectional image rescaling method based on invertible neural networks. Pan et al.~\cite{pan2022towards} achieve bidirectional rescaling of arbitrary images by joint optimization. 
% \yu{Xing et al.\cite{guo2022invertible,pan2022learning,li2021approaching,hou2018learning,zhong2022faithful,wang2022cnn} present a scale-arbitrary invertible image downscaling network. Guo et al.\cite{guo2022invertible} propose an invertible single image rescaling
% framework via steganography. Pan \cite{pan2022learning} proposes a joint optimization method. Li et al.\cite{li2021approaching} propose a flow guidance image rescaling network. Hou et al\cite{hou2018learning} propose a deep feature consistency network.
% Zhong et al.\cite{zhong2022faithful} propose the GRAIN framework for generating faithful high-resolution images. Wang et al.\cite{wang2022cnn} develop the novel CNN-based rescaling 
% algorithm.} 
The above methods elaborately design the model architectures to reconstruct better images. Our method differs from them by directly optimizing the primeval input of the IR task under the supervision of the ground-truth HR image.

\subsection{Image Super-Resolution}
Image super-resolution (SR) is a widely-used image upscaling method that refers to recovering HR images from existing LR images. It is widely used in many applications, such as object
detection~\cite{girshick2015region}, face recognition~\cite{mudunuri2015low}, medical imaging~\cite{huang2017simultaneous}, and surveillance security~\cite{rasti2016convolutional}. Existing SR methods can be divided into three categories: interpolation-based, reconstruction-based, and learning-based methods. With the help of deep learning techniques, some SR methods achieve advanced effects on learning powerful prior information~\cite{dong2015image,zhang2018image,zhang2018residual,guo2020closed,ledig2017photo,li2019feedback,guo2020hierarchical}. Ledig et al.~\cite{ledig2017photo} first propose SRGAN using Generative Adversarial Nets (GAN)~\cite{goodfellow2020generative} to solve the over-smoothed problem of the SR task. Zhang et al.~\cite{zhang2018image} combine the channel attention mechanism with SR to improve the representation ability of the model. Li et al.~\cite{li2019feedback} propose SRFBN to refine low-level information using high-level ones through feedback connections and learn better representations.
Instead of manually designing SR models, many efforts~\cite{guo2020hierarchical} have been made to automatically devise effective SR architectures using the neural architecture search (NAS) techniques~\cite{guo2019auto,chen2021contrastive,guo2022pareto}.
Nevertheless, most existing SR models contain a large number of parameters and come with very high computational cost. To address this, some recent work exploit model compression techniques~\cite{zhuang2018discrimination,liu2021discrimination,liu2021conditional} to obtain compact SR models~\cite{guo2022towards}.
Compared with SR methods that consider a fixed downscaling scheme, existing IR methods often achieve significantly better reconstruction performance thanks to the learned downscaled representations.
% However, the SR task suffers from an ill-posed problem in which a single LR image may correspond to multiple HR images, resulting in limited improvements.

\subsection{Collaborative Example}
Generating collaborative examples~\cite{li2022collaborative} and adversarial examples~\cite{madry2017towards,guo2022improving} are a set of opposite processes. The adversarial attack refers to the process of generating an adversarial example by applying a minor perturbation to the original input and causing the model to make an incorrect inference. The gradient-based attack method increases the prediction loss by updating the example along the positive direction of the gradient. In contrast to the adversarial example, the collaborative example aims to improve the robustness of the model by updating the example along the negative direction of the gradient, which ultimately decreases the prediction loss of the model. Inspired by the collaborative example, we propose a hierarchical collaborative example generation algorithm to generate downscaled representation optimal for the upscaling model under the supervision of the HR image.

%-------------------------------------------------------------------------

\begin{figure}[tp]
\centering 
\hspace{-3mm}
\includegraphics[width=\linewidth]{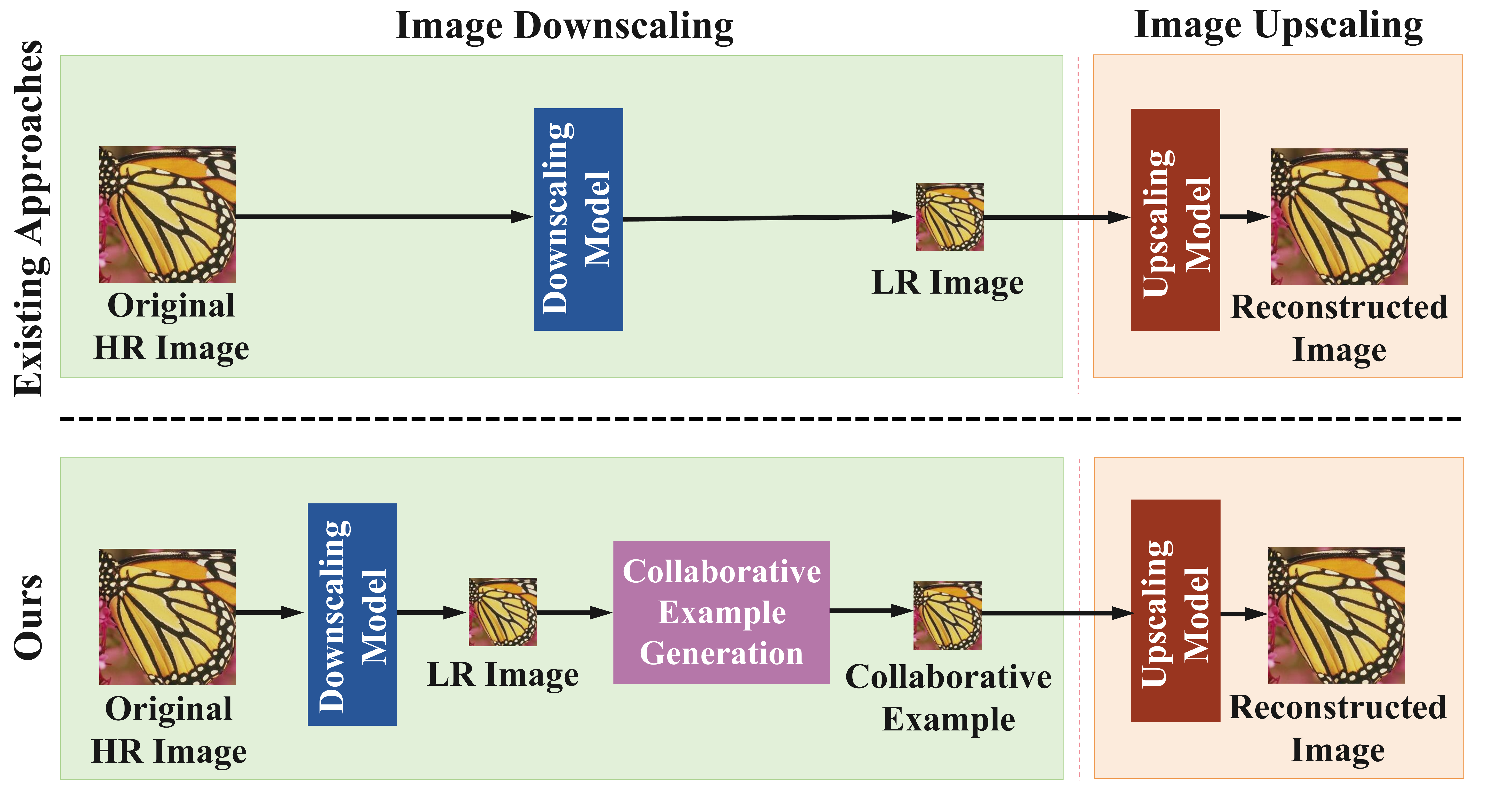} 
\vspace{5 pt}
    \caption{Comparison between existing image rescaling methods and the proposed HCD method. We additionally generate the collaborative LR examples to improve the downscaled representation while keeping the upscaling process unchanged.
    }
\label{app}     
% \vspace{-10 pt}
\end{figure}

     \begin{figure*}[htbp]
     	\centering
     	\includegraphics[width=1\linewidth]{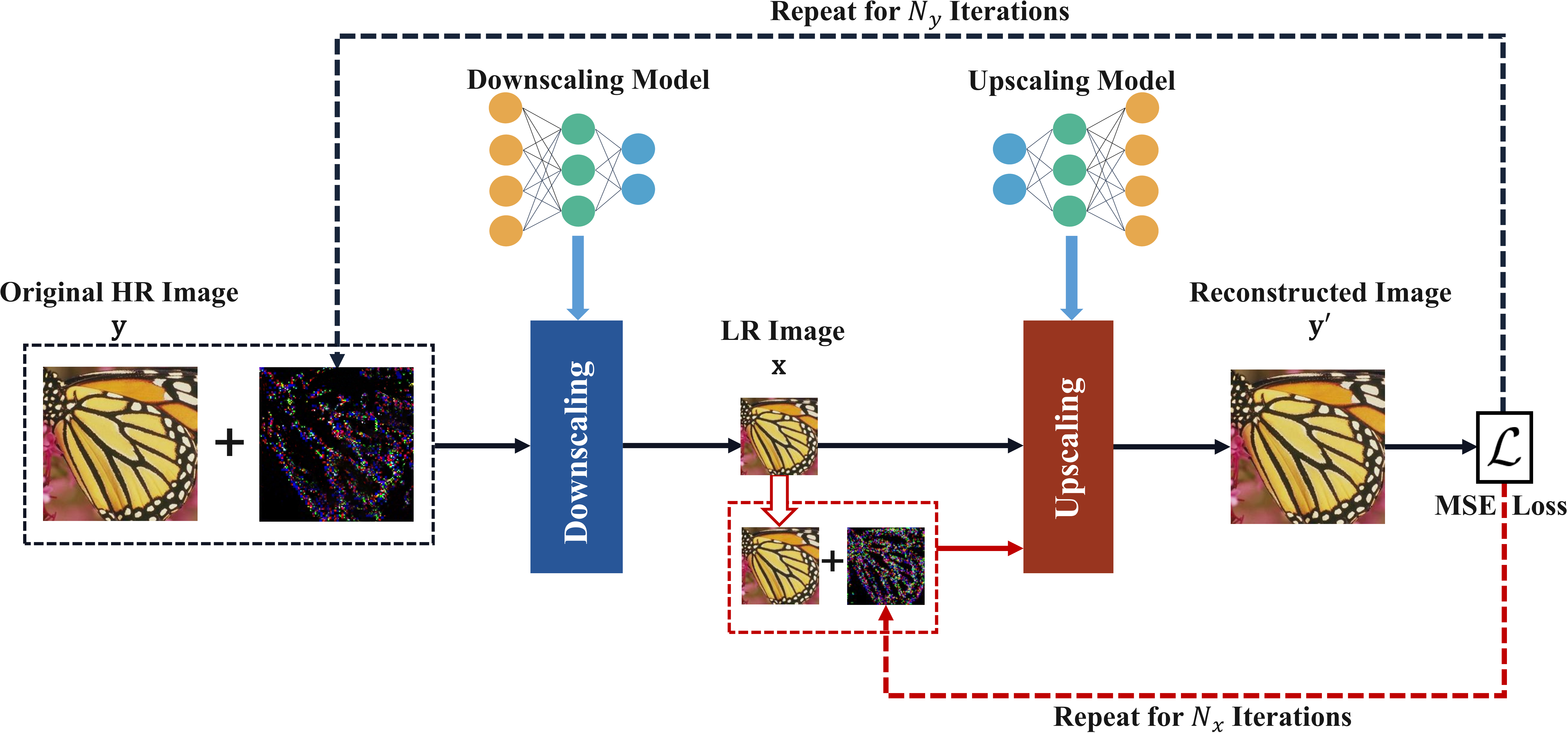}
     	% \caption{The architecture of our HCD. We sequentially optimize the original HR image $y$ and the downscaled image $x$ obtained from $y$.}
     	\vspace{1pt}
     	\caption{The proposed Hierarchical Collaborative Downscaling (HCD) scheme consists of two processes, including the collaborative HR example generation (marked as black lines) and the collaborative LR example generation (marked as red lines). We first iteratively optimize the perturbation on the original HR image to generate the collaborative HR examples. Then, we obtain the downscaled image and generate collaborative LR examples for it. In the end, we can get a better high-resolution image from the better downscaled representation.}
     	\label{architecture}
     \end{figure*}

\section{Collaborative Image Downscaling}
\label{sec:method}
    % The main idea of this paper mainly comes from the proposal of adversarial samples and collaborative samples in adversarial training. Adversarial samples refer to the sample formed by adding subtle noise to the sample, which is often not easy to be perceived visually, but will make the model show high losses, and is often used to train the model to defend against adversarial attack. Collaborative samples are defined as those that exhibit less predictive loss than clean samples but are perceptually similar to adversarial and clean samples. \par

    % In this paper, we seek to improve the performance of image rescaling by learning a better downscaled representation.
    
    % In this paper, we propose a plug-and-play method Hierarchical Collaborative Downscaling (HCD) to improve the performance of image rescaling. To this end, we first discuss the importance of downscaled representation in Section \Cref{3.1}. Then, we propose a hierarchical collaborative example generation algorithm for image rescaling task, which can improve the upscaling performance from the downscaled images.
    % We show the overall architecture of HCD in Fig.~\Cref{architecture}.
    % In general, our method consists of two stages. Firstly, we add some perturbations to the ground-truth HR images and feed them into the downscaling model to generate the corresponding LR images. Then, we add some perturbations to these LR images and feed them into the upscaling model to reconstructed HR images. 

    In this paper, we seek to directly learn a better downscaled representation rather than the down-/upscaling models to improve the performance of reconstructed images. In \Cref{3.1}, we first discuss the importance of downscaled representation in image reconstruction tasks. Besides learning a good model, directly optimizing the downscaled representation (\textcolor{mypurple}{purple box} in~\Cref{app}) is an effective way to improve performance. In \Cref{3.2}, we further extend this idea and propose a Hierarchical Collaborative Downscaling (HCD) method that optimizes the representations in both high-resolution (HR) and low-resolution (LR) domains to obtain a better downscaled representation.
    % focuses on the input at different stages of the image rescaling task, gradually optimizing the LR image. 
    The overview of our HCD method is shown in ~\Cref{architecture}.
    %We show the overall architecture of HCD in ~\Cref{architecture}.

   \subsection{Downscaled Representation Matters}
   \label{3.1}

    Existing image rescaling (IR) methods~\cite{xiao2020invertible,sun2020learned,pan2022towards} essentially learn the optimal downscaled representation by jointly training the downscaling and upscaling models to reconstruct the original HR images. 
    Compared with super-resolution methods that consider a fixed downscaling kernel, e.g., bicubic, IR methods often yield significantly better results thanks to the improved downscaled representation. For example, as shown in ~\Cref{constrast}, a recent rescaling method HCFlow~\cite{Liang2021HierarchicalCF} outperforms a strong baseline by a large margin of 5.27 dB. Such a large performance gap indicates the importance of a good downscaled representation in image reconstruction tasks.

    In order to improve the downscaled representation, all the existing rescaling methods learn a downscaling method to produce the LR images. Besides training the model, one can also directly optimize the downscaled representation itself. From this perspective, a popular approach is adversarial attack~\cite{qiu2019review} that learns the optimal perturbations on data without changing the model parameters. As shown by the \textcolor{red}{red box} in~\Cref{col vs adv}, the adversarial attacks against image rescaling models greatly hamper the reconstruction performance in terms of both PSNR and visual quality. Nevertheless, we seek to improve the performance instead of degrading it. To address this issue, a simple and intuitive way is to generate collaborative examples by considering an opposite training objective to adversarial attack.
    Specifically, we seek to generate collaborative examples in the LR domain by minimizing the reconstruction loss. As shown by the \textcolor{blue}{blue box}, we can simultaneously improve PSNR and obtain visually plausible details in the reconstructed HR image. We highlight that the PSNR improvement of 0.93 dB is significant in image reconstruction tasks and generating collaborative examples provides us with new insights.
    
    % From this perspective, the quality of the downscaling representation affects the final performance of the upscaling model. Therefore, we propose a novel method that directly optimizes the downscaled representation for the upscaling model instead of updating the model parameters. A typical technique to directly optimize the input is adversarial attack.
    % Adversarial attack iteratively perturbates the input image itself, generating adversarial example~\cite{madry2017towards} to mislead the model. Following~\cite{madry2017towards}, based on the popular IR model HCFlow~\cite{Liang2021HierarchicalCF}, we generate adversarial examples of LR images in the gradient ascent direction and feed them into the upscaling model. As shown in the red box in ~\Cref{col vs adv}, we show the reconstruction effect of the HCFlow~\cite{Liang2021HierarchicalCF} on the adversarial examples. The results show that the PSNR has dropped by 0.93db. Therefore, it is natural to think that a collaborative example can be generated in the opposite direction of generating the adversarial example so that the upscaling model can get a better solution by getting a better LR image. As shown in the blue box in ~\Cref{col vs adv}, the collaborative example improves the PSNR of the reconstructed images when the model parameters are kept constant. 
    
 % From this perspective, the downscaled representation proves its importance in IR. 
   
   \subsection{Hierarchical Collaborative Downscaling} 
    \label{3.2}
    In this part, we further extend the above idea and propose a novel Hierarchical Collaborative Downscaling (HCD) scheme to improve the downscaled representation, as shown in~\Cref{architecture}. It is worth noting that, since LR images are directly obtained from the corresponding HR images, one can also improve the LR representation based on a better example in the HR domain. In this way, it becomes possible to obtain a better downscaled representation by generating collaborative examples in both the LR and HR domains. Essentially, jointly learning both collaborative examples can be regarded as a bi-level optimization problem. Before discussing it, we first start with the generation of collaborative examples in each domain separately.

    \paragraph{Generating collaborative LR examples.}
    % We can directly learn the upscaling process to improve the downscaled representations. We calculate the gradient of MSE loss between the output of the upscaling model with the original downscaled LR image $x$ as input and the real HR image $y$, and update $x$ in the direction of gradient descent. The objective is described as~\cref{eq1}. 
    We first focus on the LR domain to generate collaborative examples.
    Basically, the standard image rescaling scheme seeks to map an HR image $y$ to a downscaled image $x$ and then upscale it to obtain a reconstructed HR image $y'$. 
    To improve the reconstruction performance, we fix the model parameters and directly learn the optimal perturbation $\delta_x$ to improve the downscaled representation $x$. Let $\mathcal{L}$ be the reconstruction loss, $f(\cdot)$, and $g(\cdot)$ denote the upscaling and downscaling model, respectively. Following ~\cite{madry2017towards}, we constrain the perturbation within a p-norm epsilon ball to avoid significantly changing the visual content, i.e., $\|\delta_x\|_p \leq \epsilon$. Formally, the perturbations can be obtained by minimizing $\mathcal{L}$:
    \begin{equation}
    \begin{aligned}
        \delta_x &= \argmin_{\|\delta_x\|_p \leq \epsilon} \mathcal{L}(f(x+\delta_x), y).
    \label{eq1}
    \end{aligned}
    \end{equation}  
    % where $f(\cdot)$ is upscaling model and $g(\cdot)$ is the downscaling model. 
    Thus, the resultant downscaled representation can be obtained by summing up the original LR image and the learned perturbation via $x := x + \delta_x$.
    % We can simply learn $\delta_x$ such that the reconstruction loss can be minimized after $x+\delta_x$ is input to the upscaling model. The improved downscaled representation leads to better reconstruction results. Since $x$ is obtained by $g(y)$, We can also get a better representation of $x$ by improving $y$ i.e. learn $\delta_y$. Inspired by this, we propose a Hierarchical Collaborative Downscaling (HCD) method that sequentially optimizes $y$ and $x$ obtained from $y$. 

    % \paragraph{Definition 2 (HR collaborative examples)}
    \paragraph{Generating collaborative HR examples.}
    Since the LR image $x$ can be obtained by downscaling the HR image $y$, we can also generate a collaborative HR example to obtain a better downscaled representation. 
    % In this way, the HR collaborative example enables the generated downscaled representation to recover the original HR image. 
    Similarly, we add additional perturbation $\delta_y$ to the input HR image $y$ and optimize it by minimizing the reconstruction loss $\mathcal{L}$. Unlike~\Cref{eq1}, we discard the perturbation in the LR domain and sequentially perform downscaling $g(\cdot)$ and upscaling $f(\cdot)$ to obtain the reconstructed image $y'$. Here, we consider the same constraint w.r.t. the epsilon ball.
    Thus, the perturbation in the HR domain can be obtained by
    \begin{gather}
        \delta_y = \mathop{\arg\min}_{\|\delta_y\|_p \leq \epsilon} \mathcal{L}(f(g(y + \delta_y)), y).  
    \end{gather}
    % q: delta y OR y^* ???

    As illustrated above, we can obtain a better downscaled representation by generating either collaborative LR examples or collaborative HR examples. This motivates us to explore whether it is possible to further improve the downscaled representation if we combine them together.
    Inspired by this, we develop a hierarchical scheme that jointly optimizes the representations in both LR and HR domains.
    Due to the dependence of the LR image on the corresponding HR image, we consider a bi-level optimization problem to obtain the optimal LR perturbation:
    \begin{gather}
        \delta_x = \mathop{\arg\min}_{\|\delta_x\|_{p} \leq \epsilon} \mathcal{L} (f (g(y + \delta_y) + \delta_x), y),  \notag \\
        {\rm s.t.}~~ \delta_y = \mathop{\arg\min}_{\|\delta_y\|_{p} \leq \epsilon} \mathcal{L}(f(g(y + \delta_y)), y).
        \label{bilevel}
    \end{gather}
    To solve this problem, we propose to alternatively generate collaborative examples in HR and LR domains.
    As shown in Algorithm~\cref{Algorithm1}, we first generate collaborative HR examples and downscale them to get the downscaled images. Then, we further generate collaborative examples w.r.t. the latest LR images to obtain the resultant downscaled representation. Following the optimization process of Project Gradient Descent (PGD) attack~\cite{madry2017towards}, we optimize the perturbations $\delta_x$ and $\delta_y$ in an iterative manner. For simplicity, we force both collaborative example generation processes to share the same number of iterations to perform gradient descent $N_x = N_y$.  In this paper, we consider $\ell_2$-norm to build the epsilon ball, i.e., $p=2$.

    \floatname{algorithm}{Algorithm}  
    \renewcommand{\algorithmicrequire}{\textbf{Input:}}  
    \renewcommand{\algorithmicensure}{\textbf{Output:}}  
    \begin{algorithm}[tb]         
        \caption{Learning scheme of \textbf{Hierarchical Collaborative Downscaling (HCD)}. We present a hierarchical learning scheme that sequentially generates the collaborative HR examples and collaborative LR examples.}  
        \label{Algorithm1}
        \begin{algorithmic}[1]
            \Require HR image $y$, the downscaling model $g(\cdot)$, the upscaling model $f(\cdot)$, number of iterations $N_y$ and $N_x$, perturbation budget $\epsilon$, step size $\alpha$, the clipping function to constrain the input within feasible range $Clip\{\cdot\}$.
            \Ensure Reconstructed high-resolution image $y'$.
            \State Initialize the perturbations $\delta_y$ and $\delta_x$
            \State \emph{// Generate collaborative HR examples}
            \For{$t=1$ \text{to} $N_y$}
                %\State Input $y$ to $f$ and $g$ to obtain the gradient
                %\Statex \qquad \qquad $g = \nabla_{\delta_y}\mathcal{L}(f(g(y + \delta_y)),y)$
                \State Compute gradient via $g = \nabla_{\delta_y}\mathcal{L}(f(g(y + \delta_y)),y)$
                %\State Update $\delta_y$ by applying the gradients calculated focused on the $L_2$ norm for the minimization step as 
                \State Update $\delta_y$ via $\delta_y \gets Clip \{\delta_y - \alpha * \frac{g}{\|g\|} , \epsilon\}$
                %\Statex \qquad \qquad $\delta_y \gets Clip \{\delta_y - \alpha * \frac{g}{\|g\|} , \epsilon\}$
                %\Statex \qquad \qquad $\delta_y \gets Clip_{\epsilon} (\delta_y - \alpha * \frac{g}{\|g\|}) $
            \EndFor
            
            \State Obtain the collaborative HR example: $y = y + \delta_y$             
            \State Compute low resolution images: $x = g(y)$
            \State \emph{// Generate collaborative LR examples}
            \For{$t = 1$ \text{to} $N_x$} 
                %\State Input $x$ to $f$ to obtain the gradient
                \State Obtain gradient via $g = \nabla_{\delta_x} \mathcal{L}(f(x + {\delta_x}) , y)$
                %\Statex \qquad \qquad $g = \nabla_{\delta_x} \mathcal{L}(f(x + {\delta_x}) , y)$
                %\State Update $\delta_x$ by applying the gradients calculated focused on the $L_2$ norm for the minimization step as 
                %\Statex \qquad \qquad ${\delta_x} \gets Clip_{\epsilon}({\delta_x} - \alpha * \frac{g}{\|g\|})$
                \State Update $\delta_x$ via ${\delta_x} \gets Clip\{{\delta_x} - \alpha * \frac{g}{\|g\|},\epsilon\}$
            \EndFor

            \State Obtain the collaborative LR example: $x=x + {\delta_x}$
            \State Obtain the reconstructed image: $y' = f(x)$
        \end{algorithmic}  
    \end{algorithm}
%-------------------------------------------------------------------------
\section{Experiments}
\label{sec:experiment}
In the experiments, we evaluate the effectiveness of HCD based on two popular image rescaling methods, including IRN~\cite{xiao2020invertible} and HCFlow~\cite{Liang2021HierarchicalCF}.
% on various common benchmark datasets. 
We first describe the %\red{experimental}%
implementation details in %Section~\Cref{sec:4.1}. 
\Cref{sec:4.1}. 
Then, we compare our method with current advanced methods in %Section~\Cref{sec:4.2} 
\Cref{sec:4.2} on informative quantitative and qualitative analyses.
% \deng{Besides, we} visualize the perturbation in %Section~\Cref{sec:4.3} 
% \deng{\Cref{sec:4.3}} to help intuitively understand the mechanism of our HCD.
Finally, we conduct abundant ablation studies and raise further discussions in %Section~\Cref{sec:4.4}.} 
\Cref{sec:4.4}. Both our source code and all the collaborative examples along with the corresponding reconstruction images will be released soon.

    \begin{table*}[!ht]
    	\centering 
    	\begin{tabular}{c|c|c|c|c|c } 
    		\hline
    		{Method  } & Set5 & Set14 & BSD100 & Urban100 & DIV2K \\ 
    		\hline \hline
    		{Bicubic \& Bicubic } &28.42 / 0.8104 &26.00 / 0.7027 &25.96 / 0.6675 &23.14 / 0.6577  &26.66 / 0.8521   \\
    		{Bicubic \& SRCNN  }\cite{Dong2014LearningAD} &30.48 / 0.8628 &27.50 / 0.7513 &26.90 / 0.7101  & 24.52 / 0.7221  &–   \\
    		{Bicubic \& RDN }\cite{zhang2018residual} &32.47 / 0.8990 &28.81 / 0.7871 &27.72 / 0.7419 &26.61 / 0.8028 &–   \\
    		{Bicubic \& EDSR }\cite{Lim2017EnhancedDR} & 32.62 / 0.8984 &28.94 / 0.7901  &27.79 / 0.7437 &26.86 / 0.8080 &29.38 / 0.9032   \\
    		{Bicubic \& RCAN}\cite{zhang2018image} & 32.63 / 0.9002&28.87 / 0.7889 &27.77 / 0.7436  &26.82 / 0.8087 & 30.77 / 0.8460  \\
    		{Bicubic \& RFANet}\cite{Liu2020ResidualFA} &32.66 / 0.9004 & 28.88 / 0.7894& 27.79 / 0.7442 &26.92 / 0.8112 & 31.41 / 0.9187  \\
    		{Bicubic \& RRDB }\cite{Wang2018ESRGANES} & 32.74 / 0.9012 &29.00 / 0.7915 &27.84 / 0.7455 &27.03 / 0.8152 &30.92 / 0.8486   \\
    		{Bicubic \& SwinIR}\cite{liang2021swinir}&  32.72 / 0.9021 &28.94 / 0.7914 &27.83 / 0.7459 &27.07 / 0.8164 & –\\
    		
    		{TAD \& TAU }\cite{kim2018task} &31.81 / – &28.63 / – &28.51 / – &26.63 / – &  31.16 / – \\
    		{CAR \& EDSR }\cite{sun2020learned} & 33.88 / 0.9174  & 30.31 / 0.8382&29.15 / 0.8001 &29.28 / 0.8711 &32.82 / 0.8837   \\
    		
    		\hline
    		IRN\cite{xiao2020invertible}  & 36.19 / 0.9451 &  32.67 / 0.9015 & 31.64 / 0.8826 & 31.41 / 0.9157 & 35.07 / 0.9318\\

    		%  5 &   36.44 / 0.9471 & 33.05 / 0.9055 & 31.87 / 0.8853  & 31.56 / 0.9164 & 35.16 / 0.9321  \\ 
    		%  10 &  36.55 / 0.9482 &  33.14 / 0.9068 & 31.99 / 0.8879  & 31.70 / 0.9174 & 35.17 / 0.9322  \\ 
    		%  20 &  36.59 / 0.9485 &  33.15 / 0.9070 & 32.06 / 0.8890  & 31.72 / 0.9176 & 35.17 / 0.9322  \\ 
    		%  30 &  36.59  / 0.9486 & 33.15 / 0.9070 & 32.07 / 0.8892  & 31.73 / 0.9176 & 35.17 / 0.9322 \\ 
    		% \hline
    		
    		%  3 + 2  &  36.36 / 0.9464 &  32.92 / 0.9035 & 31.77 / 0.8843  & 31.53 / 0.9881 & 35.11 / 0.9320  \\ 
    		%  2 + 3 &  36.41 / 0.9466 & 32.97 / 0.9043 & 31.84 / 0.8854  & 31.61 / 0.9168 & 35.13 / 0.9321   \\ 
    		%  5 + 5 &  36.49 / 0.9473 &  33.09 / 0.9059 & 31.92 / 0.8872  & 31.71 / 0.9176 & 35.17 / 0.9323   \\ 
    		%  10 + 10 & 36.56 / 0.9482 & 33.18 / 0.9073 & 32.00 / 0.8889   & 31.74 / 0.9179 & 35.19 / 0.9324  \\ 
    		IRN+HCD ($N$=15) & 36.63 / 0.9488 & 33.21 / 0.9076 & 32.03 / 0.8894  & 31.76 / 0.9180 & 35.20 / 0.9324 \\ 
    		
    		\hline
    		{HCFlow }\cite{Liang2021HierarchicalCF} &36.29 / 0.9468 & 33.02 / 0.9065 & 31.74 / 0.8864 & 31.62 / 0.9206 & 35.32 / 0.9346   \\
    		HCFlow+HCD ($N$=15)& \textbf{36.99} / \textbf{0.9506} &\textbf{33.56} / \textbf{0.9116} & \textbf{32.22} / \textbf{0.8919} &\textbf{32.00} / \textbf{0.9231} & \textbf{35.36} / \textbf{0.9352}\\
    		\hline
    		
    	\end{tabular}
    	\caption{\label{Table:Quantitative } Quantitative evaluation results (PSNR / SSIM) of image reconstruction on benchmark datasets with 4$\times$ scale. The \textbf{black} values indicate the best result. HCD significantly improves  When the number of iterations N=15, our HCD-optimized IRN and HCFlow improve PSNR and SSIM metrics on each benchmark dataset. }
%    	\vspace{-10pt}
    \end{table*}
    
    \begin{figure*}
    	\centering
    	\includegraphics[width = 0.96\linewidth]{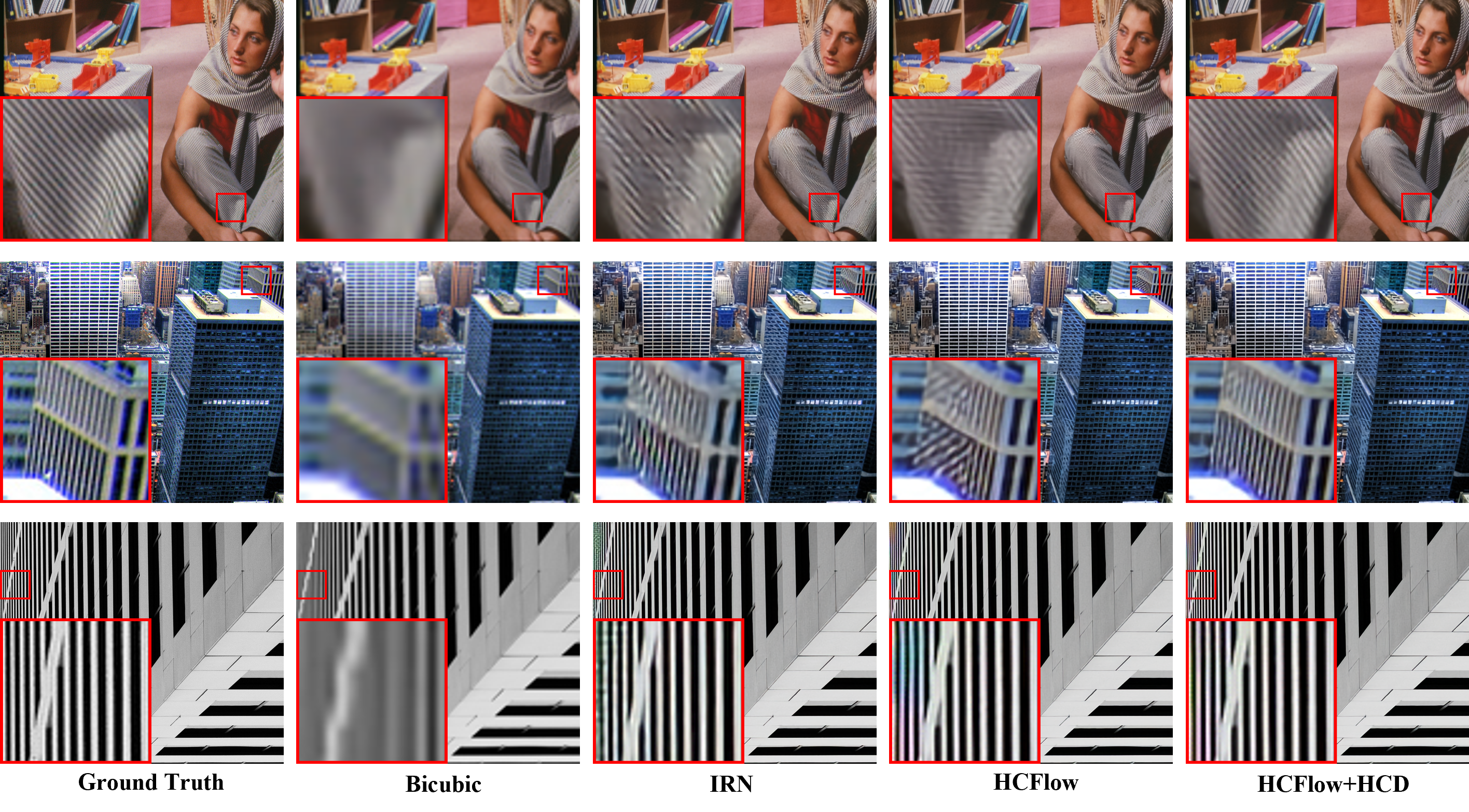}
    	\vspace{-10 pt}
    	\caption{ \label{Qualitative} Qualitative results of upscaling the 4× downscaled images. Our HCD with HCFlow is able to produce more realistic and sharper HR images compared with the baseline methods. See the appendix for more results.}
    	\vspace{-10 pt}
    \end{figure*}

\subsection{Implementation Details}
\label{sec:4.1}
We evaluate the proposed HCD on the validation set of DIV2K~\cite{agustsson2017ntire} and four standard datasets, i.e., Set5~\cite{Bevilacqua2012LowComplexitySS}, Set14~\cite{ Zeyde2010OnSI}, BSD100~\cite{Martin2001ADO}, and Urban100~\cite{Huang2015SingleIS}. We compare HCD with several state-of-the-art methods, including IRN~\cite{xiao2020invertible} and HCFlow~\cite{Liang2021HierarchicalCF}. 
Following~\cite{Lim2017EnhancedDR}, with regard to images in the YCbCr color space, we quantitatively evaluate the PSNR and SSIM~\cite{Yeo2018NeuralAC} on the Y channel of them, and test in 2× and 4× scale downscaling and reconstruction. The perturbation budget $\epsilon$ is set to 0.3 and the inner step size $\alpha$ is set to 20/255 for all experiments. By default, the iteration numbers for constructing collaborative LR images $N_x$ and HR images $N_y$ are set to the same i.e. $N_x = N_y = 15$.

\subsection{Comparisons with State-of-the-arts}
\label{sec:4.2}
     This section reports the performance of image reconstruction results on PSNR and SSIM. We consider two kinds of reconstruction methods as our baselines: IR methods and SR methods, and experiment on the best-performing IR models with few iterations. 
    % On the 2$\times$ scale downscaled and reconstruction, as Table \Cref{Table:Quantitative1 } shown, our method is used on IRN which only needs 15 iterations, and we boost the PSNR metric about 0.2-0.7 dB. On Table \Cref{Table:Quantitative },  on 4$\times$ scale downscaled and reconstruction, our method increases 0.2-0.5 dB on each benchmark dataset. When our method is used on HCFlow, the improvement reaches 0.1-0.7 dB.
    % This demonstrates that our method is suitable for these models and can achieve large improvements with little overhead. 
    It is worth mentioning that our HCD performs in the inference stage and the model parameters will not be changed during the iteration. We report the quantitative evaluation result of 4$\times$ scale and leave the results on 2$\times$ scale in the appendix due to page limit. Our HCD significantly outperforms the baseline.\par 

\begin{figure}[t]
    \centering
\includegraphics[width=0.9\linewidth]{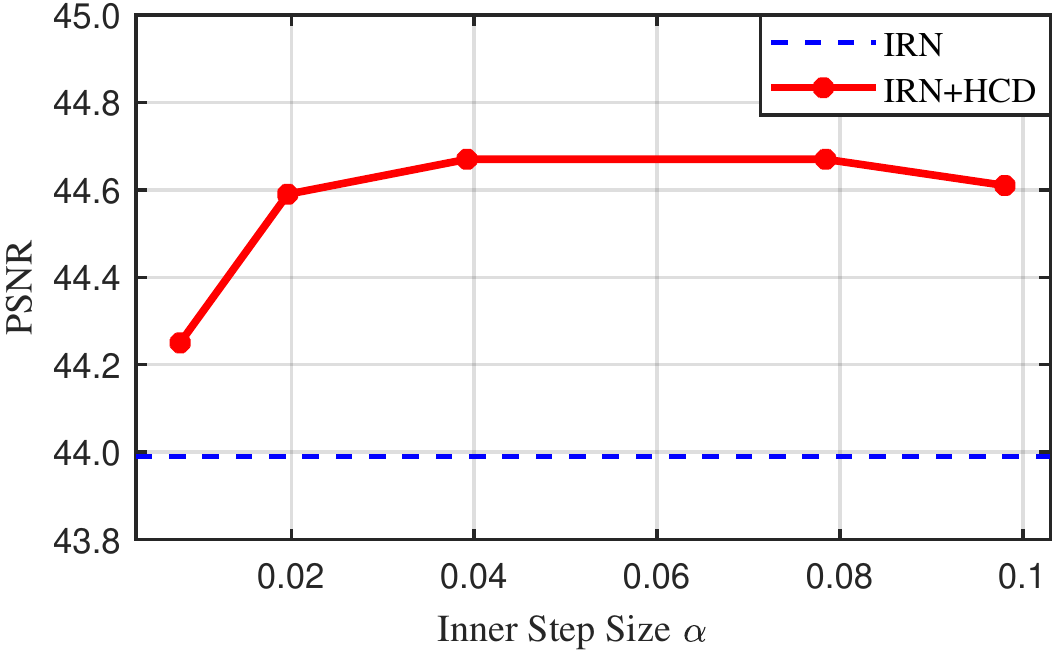}
    % \vspace{-15 pt}
    \caption{Experiment on choosing the different inner step sizes $\alpha$ on Set5 at 2$\times$ scale at the 15th iteration based on IRN. Our HCD performs well by varying the inner step size from 0.04 to 0.08.}
    \label{fig:alpha}
    % \vspace{-10 pt}
\end{figure}

\begin{table}[htbp]
    \centering 
    \setlength{\belowdisplayskip}{3pt}

        \begin{tabular}{c|c|c|c} 
        \hline
            \# Iterations $N_x$ & \# Iterations $N_y$&  PSNR & SSIM \\ \hline \hline
         15 & 0 & 44.45 & 0.9880\\
         0 & 15 & 44.37 & 0.9879\\ 
        % 30 & – & 44.46 / 0.9880 \\
       %  – & 30 & 44.41 / 0.9880\\
         15  &15  & \textbf{44.67} & \textbf{0.9886} \\

         \hline 
        \end{tabular}
    \caption{\label{Table:HR images  }The quantitative evaluation results (PSNR / SSIM) of different iteration schemes on Set5 at 2$\times$ scale based on IRN. Compared with optimizing LR or HR images separately(the second and third rows), the scheme of sequentially optimizing HR and LR images(the fourth row) increases PSNR by 0.22-0.30 dB.}
    \vspace{-10 pt}

\end{table}

    % \begin{table*}[!ht]
    % \centering 
    %     \begin{tabular}{ c|c|c|c|c|c|c } 
    %         \hline
    %         Iterations & Scale & Set5 & Set14 & BSD100 & Urban100 & DIV2K \\ 
    %         \hline \hline
    %             iters=0(CSRRevnet) & 2 $\times$ & - &  -\\ 
    %             iters=30 & 2$\times$ & - & - \\ 
    %             iters=50 & 2$\times$ & - &  -\\ 
    %             iters=70 & 2$\times$ & - & -\\ 
    %         \hline \hline
    %             iters=0(CSRRevnet) & 4$\times$ & - \\ 
    %             iters=30 & 4$\times$ & - \\ 
    %             iters=50 & 4$\times$ & - \\ 
    %             iters=70 & 4$\times$ & - \\ 
    %         \hline
    %     \end{tabular}
    % \caption{Quantitative evaluation results .}
    % \label{CSRRevnet}
    % \end{table*}
    
    % \begin{figure*}[htbp]
    %     \centering
    %     \includegraphics[scale=0.25]{image/baby.png}
    %     \caption{\label{Qualitative}Qualitative Results}
    %     \label{Qualitative}
    % \end{figure*}
    
    \paragraph{Compared methods.} To verify the effectiveness of our HCD, we investigate both IR methods and SR methods. IR methods includes HCFlow~\cite{Liang2021HierarchicalCF}, IRN~\cite{xiao2020invertible}, TAD \& TAU~\cite{kim2018task}, CAR \& EDSR~\cite{sun2020learned}. As for SR methods, we use Bicubic downsampled LR images as input, and further perform SwinIR~\cite{liang2021swinir}, RRDB~\cite{Wang2018ESRGANES}, RFANet~\cite{Liu2020ResidualFA}, RCAN~\cite{zhang2018image}, EDSR~\cite{Lim2017EnhancedDR}, RDN~\cite{zhang2018residual}, SRCNN~\cite{Dong2014LearningAD}. Compared with the above methods, our HCD has a great improvement in quantitative results and in qualitative results.

    \paragraph{Quantitative results.}  We summarize the quantitative comparison results of HCD and other methods with 4$\times$ scale in Table \ref{Table:Quantitative }. Our HCD greatly achieves better performance than previous state-of-the-art methods on PSNR and SSIM in all datasets. 
    When $N=0$, it represents the initial results of the model. 
    Compared with the original model, HCD significantly improves the reconstruction of HR images with 15 iterations. For the 4$\times$ scale reconstructed images, HCD improves by 0.1-0.7 dB compared with HCFlow and improves by 0.2-0.44 dB on IRN. Especially, on dataset Set5, HCD achieves the PSNR metric of about 36.99 dB on HCFlow. In general, our HCD significantly outperforms the baseline for PSNR and SSIM, respectively.

\paragraph{Qualitative results.} We qualitatively evaluate our HCD by demonstrating details of the upscaled images. As shown in~\Cref{Qualitative}, the results of HCD based on HCFlow achieve exhibit superior details and attractive visual quality.
In the last set of~\Cref{Qualitative}, our HCD alleviates unnatural colors in images from IRN and HCFlow.
And it produces neater lines without bothersome horizontal lines compared with IRN. 
This demonstrates that our HCD significantly outperforms HCFlow and IRN visually.\par
% Moreover, due to no additional parameters needing to be stored for our method, it will not take up more storage space than the original model. In the inference stage, we only add some minor perturbations to the LR and HR images by an iterative method, which improve the performance of some upscaling model\cite{xiao2020invertible,Liang2021HierarchicalCF}. This indicates that our method can make the server store the same amount of data, which can result in higher-quality reconstructed images.
% \subsection{Comparison between low-resolution images}
% We also evaluate the quality of LR images downscaled by our method based on IRN. We demonstrate the similarity index between our LR images and Bicubic-based LR images, and present similar visual perceptions of them on a, to show that the LR image is able to perform as well asBicubic. As shown in Table \Cref{Table:SSIM }, images downscaled by our method are extremely similar to those by Bicubic, which demonstrates the proper perception of our downscaled images.

\begin{figure}[t]
    \centering
\includegraphics[width=0.9\linewidth]{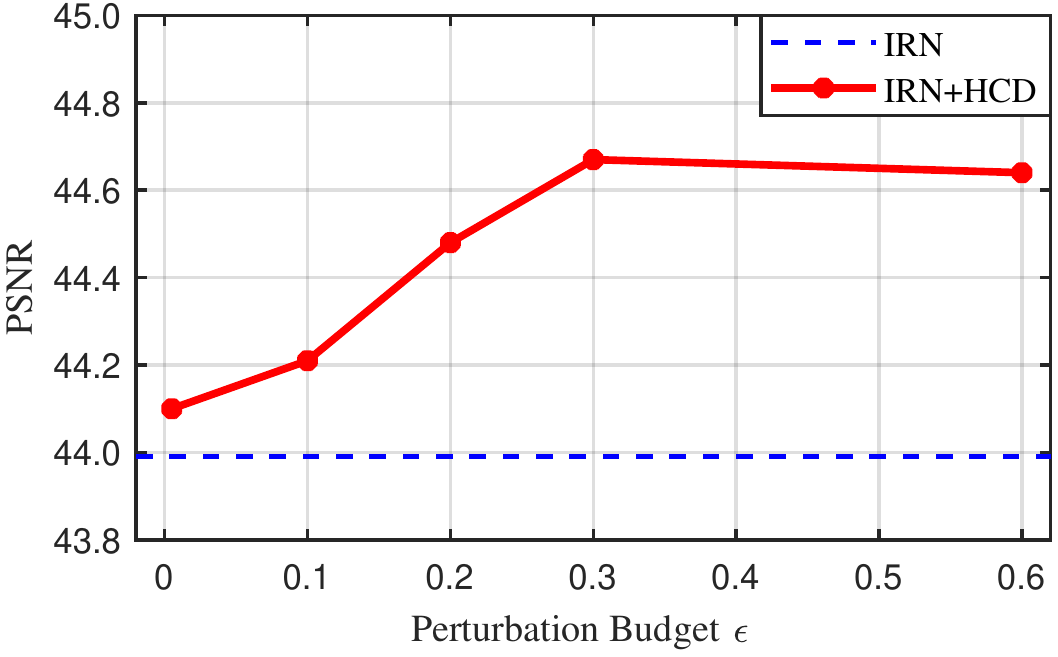}
    \caption{Experiment on choosing the different perturbation budget $\epsilon$ on Set5 at 2$\times$ scale at the 15th iteration based on IRN. The perturbation budget $\epsilon$ is expected to be as small as possible to bring less perturbation while maintaining performance. A moderate value of $\epsilon=0.3$ brings a significant improvement.}
    \label{fig:epsilon}
\end{figure}

\begin{figure*}
    \centering
    \includegraphics[width = 0.93\linewidth]{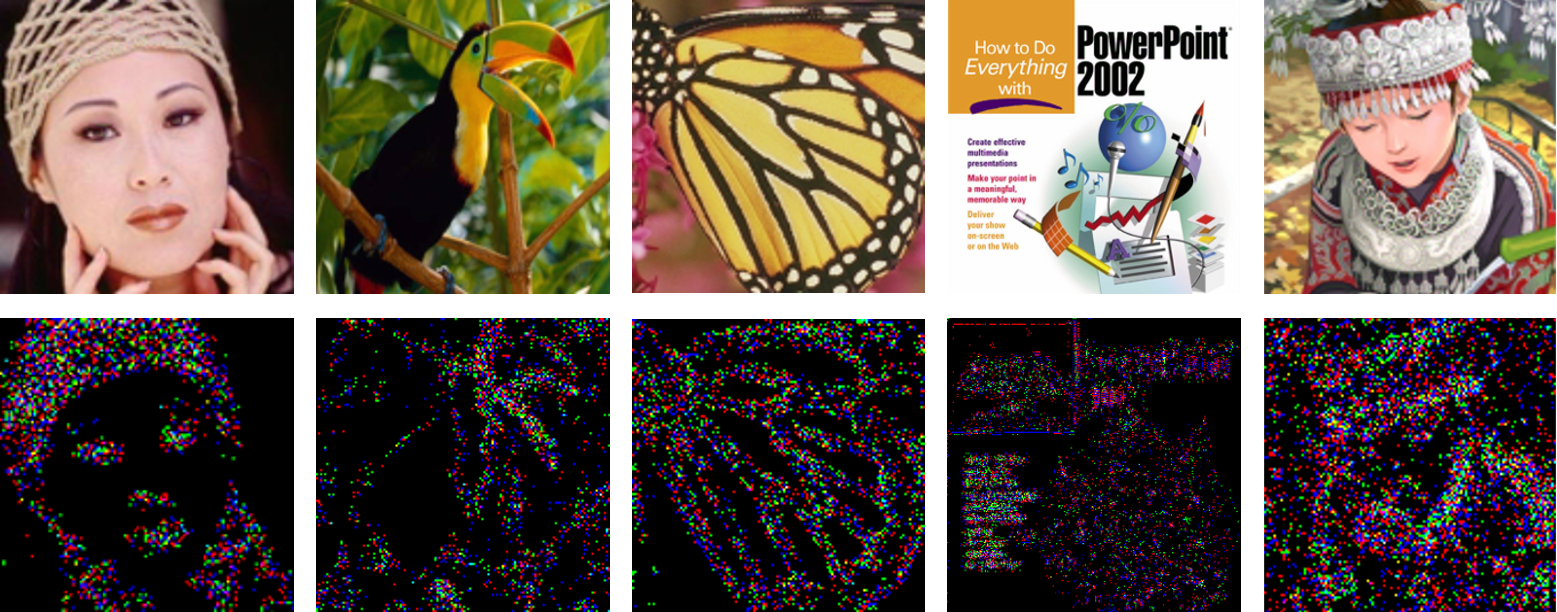}

  \caption{\label{fig: perturbation}Visualization of the generated collaborative LR examples (top) and the corresponding perturbations $\delta_x$ (bottom). The perturbations added to the LR image are mainly distributed on the contours and corners of the image. These visualization results indicate that the generated collaborative examples are able to provide more information for those high-frequency regions, which, however, are often hard to be reconstructed in the upscaling process.}
  \vspace{6 pt}
\end{figure*}

    \begin{table*}[htbp]
    \centering 
        \begin{tabular}{ c|c|c|c|c} 
        \hline
         \# Iterations $N$ &  PSNR & SSIM &  Downscaling Latency  & Upscaling Latency \\ \hline \hline
         1 & 44.10 & 0.9872 & 0.33s & \multirow{5}{*}{\textbf{0.03s}} \\
         5 &  44.52 & 0.9882 & 1.61s & \\
         10 & 44.61 & 0.9884 & 2.98s & \\
         15 & \textbf{44.67} & 0.9886 & 3.97s & \\
         20 & 44.66 & 0.9872 & 5.26s & \\

        % 1 & 44.05 / 0.9871 \\
         % 2 & 44.19 / 0.9873 \\
         % 3 & 44.32 / 0.9877 \\ 
         % 4 & 44.40 / 0.9879 \\
         % 5 & 44.42 / 0.9879 \\
         % 
         \hline 
        \end{tabular}
        
    \caption{
    Effect of the number of iterations for generating collaborative examples ($N_x = N_y = N$) on the reconstruction performance and computational cost. We report results on Set5 for $2\times$ rescaling based on IRN.
    % The quantitative evaluation results and latency of different iterations ($N_x = N_y = N$) on Set5 at 2$\times$ scale based on IRN.
    % PSNR increases with iterations and the time overhead generated by HCD only exists in the stage of generating LR images. 
    Our HCD method gradually improves the reconstruction performance when we increase the number of collaborative iterations from 1 to 15. Nevertheless, the performance improvement becomes negligible if we further increase the iteration number to $N=20$. More critically, we highlight that our method does not increase the latency of the upscaling stage, which makes it applicable in real-world scenarios.}
\label{Table:iterations}
 \vspace{-1 pt}
    \end{table*}

\subsection{Ablations and Further Discussions}
\label{sec:4.4}
In this section, we present the results of the various iteration schemes and iteration numbers, as well as the effects of inner step size $\alpha$ and perturbation budget $\epsilon$. We also conduct ablations on HCD to justify each component. Unless otherwise specified, all ablation experiments are conducted on Set5 for 2$\times$ scale based on IRN.%没到一半
% , but when we continue to increase the number of iterations to 30, we find that we can only continue to improve 0.01 dB and 0.04 dB, 

\paragraph{Effect of hierarchical collaborative learning.} 
Table \ref{Table:HR images  } reports the performance of different iteration combinations ($N_x, N_y$) based on the backbone method IRN. When the iteration number degrades to zero, we skip the collaborative example generation step for the HR or LR images, which corresponds to line 1 and line 3 in Table \ref{Table:HR images  }, respectively. When only collaborative LR examples or collaborative HR examples are used, the reconstructed image resolution is improved by 0.46 dB and 0.38 dB after 15 iterations compared to IRN. When we leverage both examples, we can improve the performance by 0.22-0.3 dB over the best results achieved by these two examples alone. These results demonstrate the effectiveness of the proposed hierarchical learning scheme, showing that the collaborative HR examples can be combined with the collaborative LR examples to boost the image reconstruction performance. 
%Since optimizing the LR images alone will quickly reach the bottleneck after a few iterations.

\paragraph{Effect of the step size $\alpha$.}In order to explore the effect of different inner step size $\alpha$, we keep the perturbation budget $\epsilon$  fixed at  0.3 and change the $\alpha$ in a range on IRN. The inference results are shown in \Cref{fig:alpha}.  Similar to the learning rate, if it is too low, the inference process will converge slowly, causing the LR and HR images we infer to be lower than expected. 
% Or we need to iterate more times to guarantee performance. 
However, since we randomly initialize the perturbations, a too-high $\alpha$ can also produce unreliable results. Experimental results show that our HCD performs well by varying the inner step size from 0.04 to 0.08.
%The proper inner step size can enhance our performance.

\paragraph{Effect of the perturbation budget $\epsilon$.}As shown in \Cref{fig:epsilon}, We show how the inference results change as the perturbation budget changes. The maximum change in the image over the course of one iteration is related to the perturbation budget. As the figure illustrates, due to the limit of the change, a low perturbation budget can not produce a very good performance.
The high perturbation budget allows for a wide range of pixel variations, resulting dramatically image changes, so it causes subpar performance. %YZB:这句话不是很明白，感觉要修改一下
We aim to achieve excellent performance improvement with minimal changes to the original images.

\paragraph{Visualization of generated perturbations on LR images.} 
In this section, we explicitly visualize the generated perturbations on the downscaled representation, i.e., $\delta_x$ in~\cref{bilevel}. 
% In the process of forming collaborative HR and LR examples, benign perturbations are added to the image in our method to produce a better downscaled representation. 
As shown in \Cref{fig: perturbation}, the perturbations are mainly distributed on the contour and corner of the image, such as the woman's eyes and the structure of a butterfly. Interestingly, these regions often contain high-frequency information that is hard to capture in the image upscaling process. We highlight that the performance improvement of our HCD method mainly stems from these collaborative perturbations.
% So that the upscaling model can produce clearer HR images, HCD can identify the region that is sensitive to the upscaling model and add the necessary information.

\paragraph{Effect of the collaborative iteration number $N$.} In Table~\ref{Table:iterations},  we choose IRN as our backbone method and study the effect of different values of collaborative iteration number $N$ on building our collaborative example.
In the default setting, we suggest $N_x = N_y = N$. We can find that the performance of IRN+HCD improves as the number of iterations increases, and only a few iterations are required to achieve a stable and satisfactory reconstruction effect. Additionally, we report the downscaling and upscaling latency and find that running 15 iterations only takes 3.97 s during the downscaling stage. Since our method does not increase the latency of the upscaling stage, which makes it is applicable in real-world scenarios.

\section{Conclusion}
In this paper, we propose a Hierarchical Collaborative Downscaling~(HCD) method for the image rescaling task.
In the first step, we generate collaborative samples for the input HR image of the downscaling model, so that it can be downscaled into a better LR representation. Then, we generate collaborative samples for the optimized LR to further improve its reconstruction performance.
Extensive experiments show, both quantitatively and qualitatively, that our HCD significantly improves the performance on top of diverse image rescaling models.

\label{sec:conclusion}
%%%%%%%%% REFERENCES

% \clearpage
{\small
\bibliographystyle{ieee_fullname}
\bibliography{egbib}
}

\end{document}